\documentclass{article}
\usepackage{spconf,amsmath,graphicx,url}

\title{End-to-End streaming keyword spotting}
\name{Raziel Alvarez, Hyun-Jin Park
	\thanks{Authors thank Patrick Violette for his help with the training infrastructure to make this work possible.}
}
\address{Google Inc.\\
\texttt{\{raziel,hjpark\}@google.com}}
\begin{document}
\ninept
\maketitle
\begin{abstract}
We present a system for keyword spotting that, except for a front-end component for feature generation, it is entirely contained in a deep neural network (DNN) model trained ``end-to-end" to predict the presence of the keyword in a stream of audio. The main contributions of this work are, first, an efficient memoized neural network topology that aims at making better use of the parameters and associated computations in the DNN by holding a memory of previous activations distributed over the depth of the DNN. The second contribution is a method to train the DNN, end-to-end, to produce the keyword spotting score. This system significantly outperforms previous approaches both in terms of quality of detection as well as size and computation.
\end{abstract}
\section{Introduction}
\label{sec:intro}

Keyword detection is like searching for a needle in a haystack: the detector must listen to continuously streaming audio, ignoring nearly all of it, yet still triggering correctly and instantly. In the last few years, with the advent of voice assistants, keyword spotting has become a common way to initiate a conversation with them (e.g. ``Ok Google", ``Alexa", or ``Hey Siri"). As the assistant use cases spread through a variety of devices, from mobile phones to home appliances and further into the internet-of-things (IoT) --many of them battery powered or with restricted computational capacity, it is important for the keyword spotting system to be both high-quality as well as computationally efficient.

Neural networks are core to the state-of-the-art keyword spotting systems. These solutions, however, are not developed as a single deep neural network (DNN). Instead, they are traditionally comprised of different subsystems, independently trained, and/or manually designed. For example, a typical system is composed by three main components: 1) a signal processing frontend, 2) an acoustic encoder, and 3) a separate decoder. Of those components, it is the last two that make use of DNNs along with a wide variety of decoding implementations. They range from traditional approaches that make use of a Hidden Markov Model (HMM) to characterize acoustic features from a DNN into both ``keyword" and ``background" (i.e. non-keyword speech and noise) classes \cite{Alexa16, Alexa17, AlexaRaw17, AlexaDelayed18, Alexa18}. Simpler derivatives of that approach perform a temporal integration computation that verifies the outputs of the acoustic model are high in the right sequence for the target keyword in order to produce a single detection likelihood score~\cite{Hotwordv1, Agc15, Cascade17, HeySiri17, AlexaCompress16}. Other recent systems make use of CTC-trained DNNs --typically recurrent neural networks (RNNs) \cite{Ctc17}, or even sequence-to-sequence trained models that rely on beam search decoding \cite{Custom17}. This last family of systems is the closest to be considered end-to-end, however they are generally too computationally complex for many embedded applications.

Optimizing independent components, however, creates added complexities and is suboptimal in quality compared to doing it jointly. Deployment also suffers due to the extra complexity, making it harder to optimize resources (e.g. processing power and memory consumption). The system described in this paper addresses those concerns by learning both the encoder and decoder components into a single deep neural network, jointly optimizing to directly produce the detection likelihood score. This system could be trained to subsume the signal processing frontend as well as in \cite{AlexaRaw17, Raw15}, but it is typically computationally costly to replace highly optimized fast Fourier transform implementations with a neural network of equivalent quality. However, it is something we consider exploring in the future. Overall, we find this system provides state-of-the-art quality across a number of audio and speech conditions compared to a traditional, non end-to-end baseline system described in \cite{Cnn15}. Moreover, the proposed system significantly reduces the resource requirements for deployment by cutting computation and size over five times compared to the baseline system.

The rest of the paper is organized as follows. In Section \ref{sec:system} we present the architecture of the keyword spotting system; in particular the two main contributions of this work: the neural network topology, and the end-to-end training methodology. Next, in Section \ref{sec:setup} we describe the experimental setup, and the results of our evaluations in Section \ref{sec:results}, where we compare against the baseline approach of \cite{Cnn15}. Finally, we conclude with a discussion of our findings in Section \ref{sec:conclusion}.

\section{End-to-End system}
\label{sec:system}

This paper proposes a new end-to-end keyword spotting system that by subsuming both the encoding and decoding components into a single neural network can be trained to produce directly an estimation (i.e. score) of the presence of a keyword in streaming audio. The following two sections cover the efficient memoized neural network topology being utilized, as well as the method to train the end-to-end neural network to directly produce the keyword spotting score.

\subsection{Efficient memoized neural network topology}
\label{svdf}


We make use of a type of neural network layer topology called SVDF (single value decomposition filter), originally introduced in~\cite{Svdf15} to approximate a fully-connected layer with a low rank approximation. As proposed in~\cite{Svdf15} and depicted in Equation \ref{eq:basic-eqn-rank-1}, the activation $a$ for each node $m$ in the rank-1 SVDF layer at a given inference step $t$ can be interpreted as performing a mix of selectivity in time ($\mathbf{\alpha}^{(m)}$) with selectivity in the feature space ($\mathbf{\beta}^{(m)}$) over a sequence of input vectors $\mathbf{x}_t = [ \mathbf{X}_{t - T}, \cdots , \mathbf{X}_{t}]$ of size $F$.

\begin{equation}
a^{(m)}_t
=
f\left( \sum_{i=0}^{T - 1} \mathbf{\alpha}^{(m)}_{i} \sum_{j=1}^F \mathbf{\beta}^{(m)}_{j} x_{(t - T + i), j} \right)
\label{eq:basic-eqn-rank-1}
\end{equation}

This is equivalent to performing, on an SVDF layer of $N$ nodes, $N\times T$ 1-D convolutions of the feature filters $\mathbf{\beta}$ (by ``sliding" each of the $N$ filters on the input feature frames, with a stride of $F$), and then filtering each of the $N$ output vectors (of size $T$) with the time filters $\mathbf{\alpha}$.

\begin{figure}
	\centering
	\includegraphics[height=180pt]{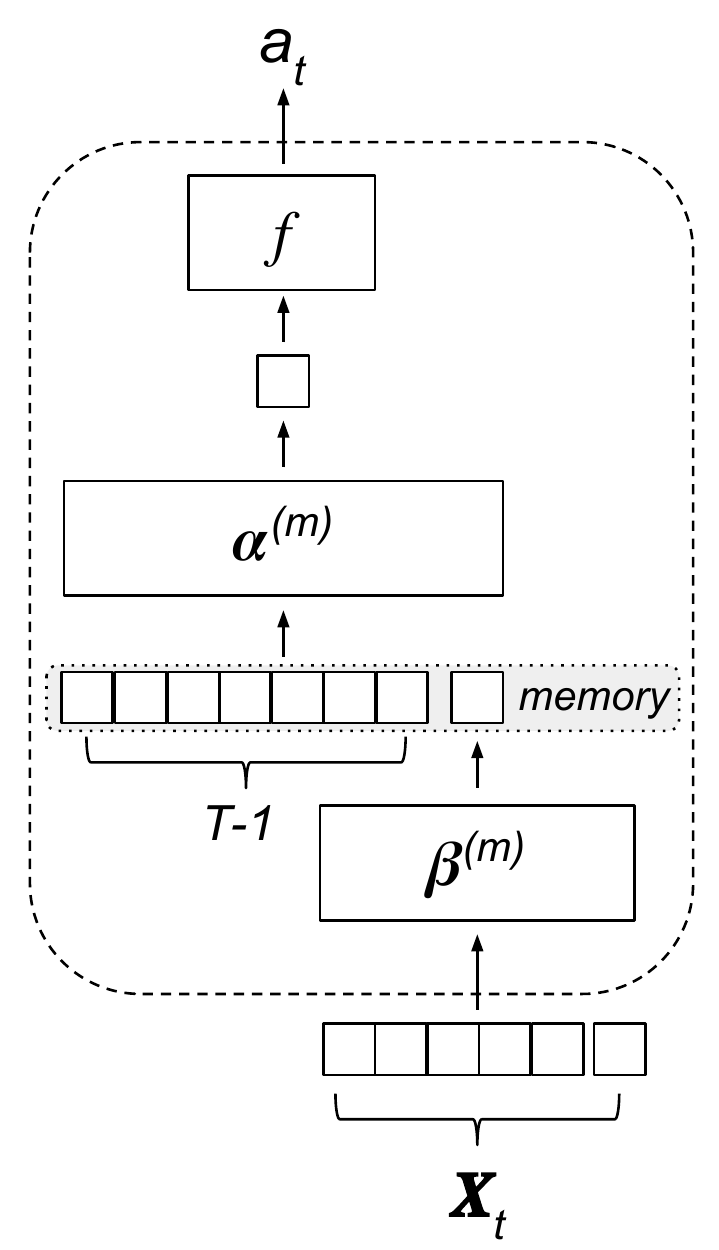}
	\caption{A single node (\emph{m}) in the SVDF layer.}
	\label{fig:svdf_node}
\end{figure}

A more general and efficient interpretation, depicted in Figure \ref{fig:svdf_node}, is that the layer is just processing a single input vector $\mathbf{X}_{t}$ at a time. Thus for each node $m$, the input $\mathbf{X}_{t}$ goes through the feature filter $\mathbf{\beta}^{(m)}$, and the resulting scalar output gets concatenated to those $T-1$ computed in previous inference steps. The memory is initialized to zero during training for the first $T$ inferences. Finally the time filter $\mathbf{\alpha}^{(m)}$ is applied to them. This is how stateful networks work, where the layer is able to memorize the past within its state. Different from other stateful approaches \cite{Fsmn15}, and typical recurrent layers, the SVDF does not recur the outputs into the state (memory), nor rewrites the entirety of the state with each inference. Instead, the memory keeps each inference's state isolated from subsequent runs, just pushing new entries and popping old ones based on the memory size $T$ configured for the layer. This also means that by stacking SVDF layers we are extending the receptive field of the network. For example, a DNN with $D$ stacked layers, each with a memory of $T$, means that the DNN is taking into account inputs as old as $\mathbf{X}_{t-D\times (T-1)}$. This approach works very well for streaming execution, like in speech, text, and other sequential processing, where we constantly process new inputs from a large, possibly infinite sequence but do not want to attend to all of it. An implementation is available at \cite{TfliteSvdf}.

This layer topology offers a number of benefits over other approaches. Compared with the convolutions used in \cite{Cnn15}, it allows finer-grained control of the number of parameters and computation, given that the SVDF is composed by several relatively small filters. This is useful when selecting a tradeoff between quality, size and computation. Moreover, because of this characteristic, the SVDF allows creating very small networks that outperform other topologies which operate at larger granularity (e.g. our first stage, always-on network has about 13K parameters~\cite{Cascade17}). The SVDF also pairs very well with linear “bottleneck” layers to significantly reduce the parameter count as in \cite{Bottleneck08, Bottleneck12}, and more recently in \cite{AlexaCompress16}. And because it allows for creating evenly sized deep networks, we can insert them throughout the network as in Figure \ref{fig:topology}. Another benefit is that due to the explicit sizing of the receptive field, it allows for a fine grained control over how much to remember from the past. This has resulted in SVDF outperforming RNN-LSTMs, which do not benefit from, and are potentially hurt by, paying attention to theoretically infinite past. It also avoids having complex logic to reset the state every few seconds as in \cite{Custom17}.

\subsection{Method to train the end-to-end neural network}
\label{training}
The goal of our end-to-end training is to optimize the network to produce the likelihood score, and to do so as precisely as possible. This means to have a high score right at the place where the last bit of the keyword is present in the streaming audio, and not before and particularly not much after (i.e. a ``spiky" behaviour is desirable). This is important since the system is bound to an operating point defined by a threshold (between $0$ and $1$) that is chosen to strike a balance between false-accepts and false-rejects, and a smooth likelihood curve would add variability to the firing point. Moreover, any time in between the true end of the keyword and the point where the score meets the threshold will become latency in the system (e.g. the ``assistant" will be slow to respond) --a common drawback of CTC-trained RNNs \cite{Ctc15} we aim to avoid.

\subsubsection{Label generation}
\label{label_gen}
\begin{figure}
	\centering
	\includegraphics[width=\columnwidth]{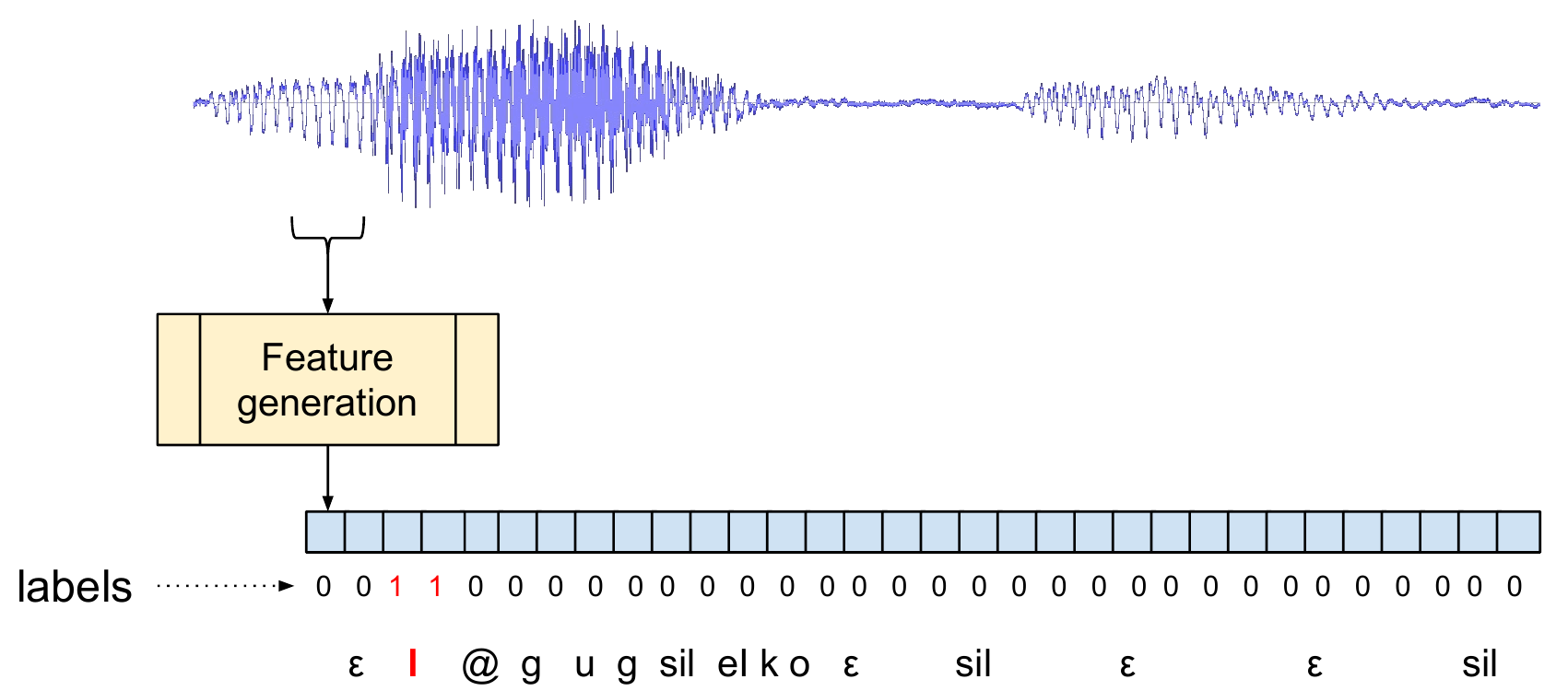}
	\caption{Input sequence generation for ``Ok google". }
	\label{fig:labeling}
\end{figure}

We generate input sequences composed of pairs \texttt{<$\mathbf{X}_{t}$,$c$>}. Where $\mathbf{X}$ is a 1D tensor corresponding to log-mel filter-bank energies produced by a front-end as in \cite{Hotwordv1, Svdf15, Cnn15}, and $c$ is the class label (one of $\{0, 1\}$). Each tensor $\mathbf{X}$ is first force-aligned from annotated audio utterances, using a large LVCSR system, to break up the components of the keyword \cite{Speech12}. For example, ``ok google" is broken into: ``ou", ``k", ``eI", ``\texttt{<silence>}", ``g", ``u", ``g", ``@", ``l". Then we assign labels of $1$ to all sequence entries, part of a true keyword utterance, that correspond to the last component of the keyword (``l" in our ``Ok google" example). All other entries are assigned a label of $0$, including those that are part of the keyword but that are not its last component. See Figure \ref{fig:labeling}. Additionally, we tweak the label generation by adding a fixed amount of entries with a label of $1$, starting from the first vector $\mathbf{X}_{t}$ corresponding to the final keyword component (e.g. ``l"). This is with the intention of balancing the amount of negative and positive examples, in the same spirit as \cite{Alexa16}. This proved important to make training stable, as otherwise the amount of negative examples overpowered the positive ones.

\subsubsection{Training recipe}
\label{recipe}
\begin{figure}
	\centering
	\includegraphics[width=\columnwidth]{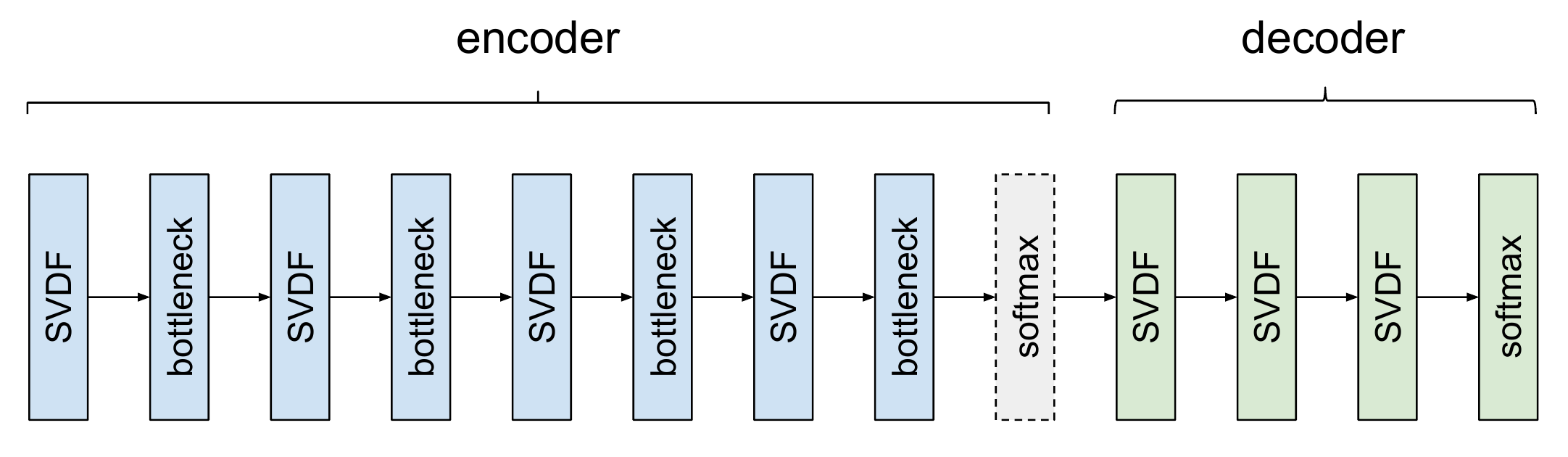}
	\caption{End-to-end topology trained to predict the keyword likelihood score. Bottleneck layers reduce parameters and computation. The intermediate softmax is used in encoder+decoder training only.}
	\label{fig:topology}
\end{figure}

The end-to-end training uses a simple frame-level cross-entropy (CE) loss that for the feature vector $\mathbf{X}_{t}$ is defined by $\lambda_{t}(\mathbf{W}) = - \log y_{c_t}(\mathbf{X}_{t}, \mathbf{W})$, where $\mathbf{W}$ are the parameters of the network, $y_{i}(\mathbf{X}_{t}, \mathbf{W})$ the $i$th output of the final softmax. Our training recipe uses asynchronous stochastic gradient descent (ASGD) to produce a single neural network that can be fed streaming input features and produce a detection score. We propose two variants of this recipe:


\textbf{Encoder+decoder}. A two stage training procedure where we first train an acoustic encoder, as in \cite{Hotwordv1,Svdf15,Cnn15}, and then a decoder from the outputs of the encoder (rather than filterbank energies) and the labels from \ref{label_gen}. We do this in a single DNN by creating a final topology that is composed of the encoder and its pre-trained parameters (including the softmax), followed by the decoder. See Figure \ref{fig:topology}. During the second stage of training the encoder's parameters are frozen, such that only the decoder is trained. This recipe is useful on models that tend to overfit to subsets of the entire training dataset.

\textbf{End-to-end}. In this option, we train the DNN end-to-end directly, with the sequences from \ref{label_gen}. The DNN may be any topology, but we use that of the encoder+decoder, except for the intermediate encoder softmax (now an unnecessary information bottleneck). See Figure \ref{fig:topology}. Similar to the encoder+decoder recipe, we can also initialize the encoder section with a pre-trained model, and use an adaptation rate $[0-1]$ to tune how much the encoder section is being adjusted (e.g. a rate of 0 is equivalent to the encoder+decoder recipe). This end-to-end pipeline, where the entirety of the topology's parameters are adjusted, tends to outperform the encoder+decoder one, particularly in smaller sized models which do not tend to overfit.

\section{Experimental setup}
\label{sec:setup}

In order to determine the effectiveness of our approach, we compare against a known keyword spotting system proposed in \cite{Cnn15}. This section describes the setups used in the results section.

\subsection{Front-end}
\label{frontend}
Both setups use the same front-end, which generates 40-dimensional log-mel filter-bank energies out of 30ms windows of streaming audio, with overlaps of 10ms. The front-end can be queried to produce a sequence of contiguous frames centered around the \emph{current} frame $\mathbf{x}_t = [ \mathbf{X}_{t - C_l}, \cdots , \mathbf{X}_{t}, \cdots, \mathbf{X}_{t + C_r}] $. Older frames are said to form the left context $C_l$, and newer frames form the right context $C_r$. Additionally, the sequences can be requested with a given stride $\sigma$.

\subsection{Baseline model setup}
\label{baseline}
Our baseline system (Baseline\_1850K) is taken from \cite{Cnn15}. It consists of a DNN trained to predict subword targets within the keywords. The input to the DNN consists of a sequence with $C_l=30$ frames of left and $C_r=10$ frames of right context; each with a stride of $\sigma=3$. The topology consists of a 1-D convolutional layer with 92 filters (of shape 8x8 and stride 8x8), followed by 3 fully-connected layers with 512 nodes and a rectified linear unit activation each. A final softmax output predicts 9 subword targets ("k" and "h" share a label for "Ok/Hey Google" detection), obtained from the same forced alignment process described in \ref{label_gen}. This results in the baseline DNN containing 1.7M parameters, and performing 1.8M multiply-accumulate operations per inference (every 30ms of streaming audio). A keyword spotting score between 0 and 1 is computed by first smoothing the posterior values, averaging them over a sliding window of the previous 100 frames with respect to the current $t$; the score is then defined as the largest product of the smoothed posteriors in the sliding window as originally proposed in \cite{Agc15}.

\subsection{End-to-end model setup}
\label{end2end}
The end-to-end system (prefix E2E) uses the DNN topology depicted in Figure \ref{fig:topology}, and all SVDF layers are of $rank=1$. We present results with 3 distinct size configurations (infixes 700K, 318K, and 40K) each representing the approximate number of parameters, and the 2 training recipe variants (suffixes 1stage and 2stage) corresponding to end-to-end and encoder+decoder respectively, as described in \ref{recipe}. The input to all DNNs consists of a sequence with $C_l=1$ frame of left and $C_r=1$ frame of right context; each with a stride of $\sigma=2$. More specifically, the E2E\_700K model uses $N=1280$ nodes in the first 4 SVDF layers, each with a memory $T=8$, with intermediate bottleneck layers each of size $64$; the following 3 SVDF layers have $N=32$ nodes, each with a memory $T=32$. This model performs 350K multiply-accumulate operations per inference (every 20ms of streaming audio). The E2E\_318K model uses $N=576$ nodes in the first 4 SVDF layers, each with a memory $T=8$, with intermediate bottleneck layers each of size $64$; the remainder layers are the same as E2E\_700K. This model performs 159K multiply-accumulate operations per inference. Finally, the E2E\_40K model uses $N=96$ nodes in the first 4 SVDF layers, each with a memory $T=8$, with intermediate bottleneck layers each of size $32$; the remainder layers are the same as the other two models. This model performs 20K multiply-accumulate operations per inference.

\subsection{Dataset}
\label{ssec:data}
The training data for all experiments consists of 1 million anonymized hand-transcribed utterances of the keywords ``Ok Google" and ``Hey Google", with an even distribution. To improve robustness, we create ``multi-style" training data by synthetically distorting the utterances, simulating the effect of background noise and reverberation. 8 distorted utterances are created for each original utterance; noise samples used in this process are extracted from environmental recordings of everyday events, music, and YouTube videos. Results are reported on four sets representative of various environmental conditions: \emph{Clean non-accented} contains 170K non-accented English utterances of the keywords in ``clean" conditions, plus 64K samples without the keywords (1K hours); \emph{Clean accented} has 153K English utterances of the keywords with Australian, British, and Indian accents (also in ``clean" conditions), plus 64K samples without the keywords (1K hours); \emph{High pitched} has 1K high pitched utterances of the keywords, and 64K samples without them (1K hours); \emph{Query logs} contains 110K keyword and 21K non-keyword utterances, collected from anonymized voice search queries. This last set contains background noises from real usage conditions.

\section{Results}
\label{sec:results}

\begin{figure}[htb]
	\begin{minipage}[b]{\linewidth}
		\centering
		\centerline{\includegraphics[height=137pt]{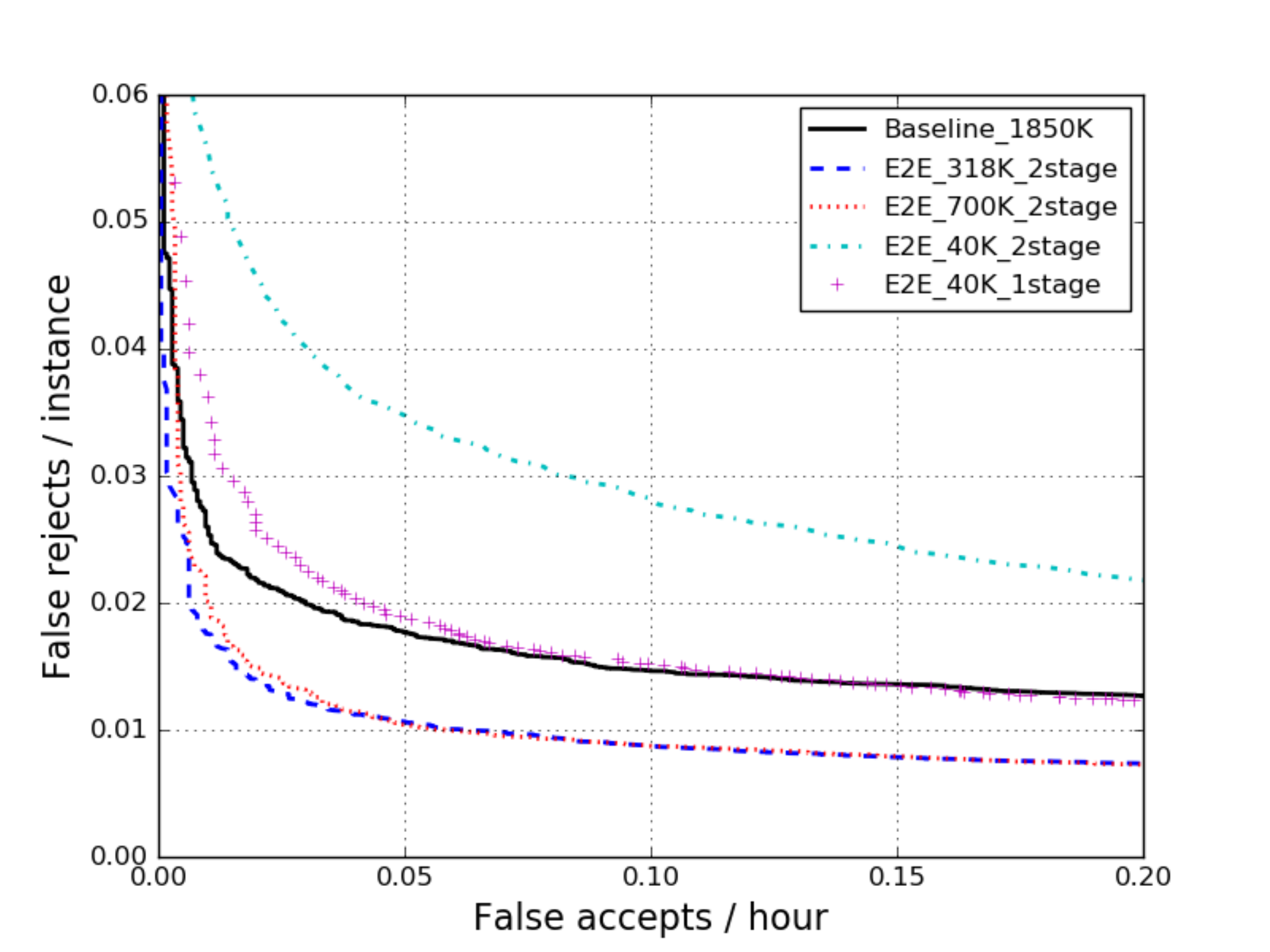}}
		\centerline{(a) Clean non-accented}\medskip
	\end{minipage}
	\hfill
	\begin{minipage}[b]{\linewidth}
		\centering
		\centerline{\includegraphics[height=137pt]{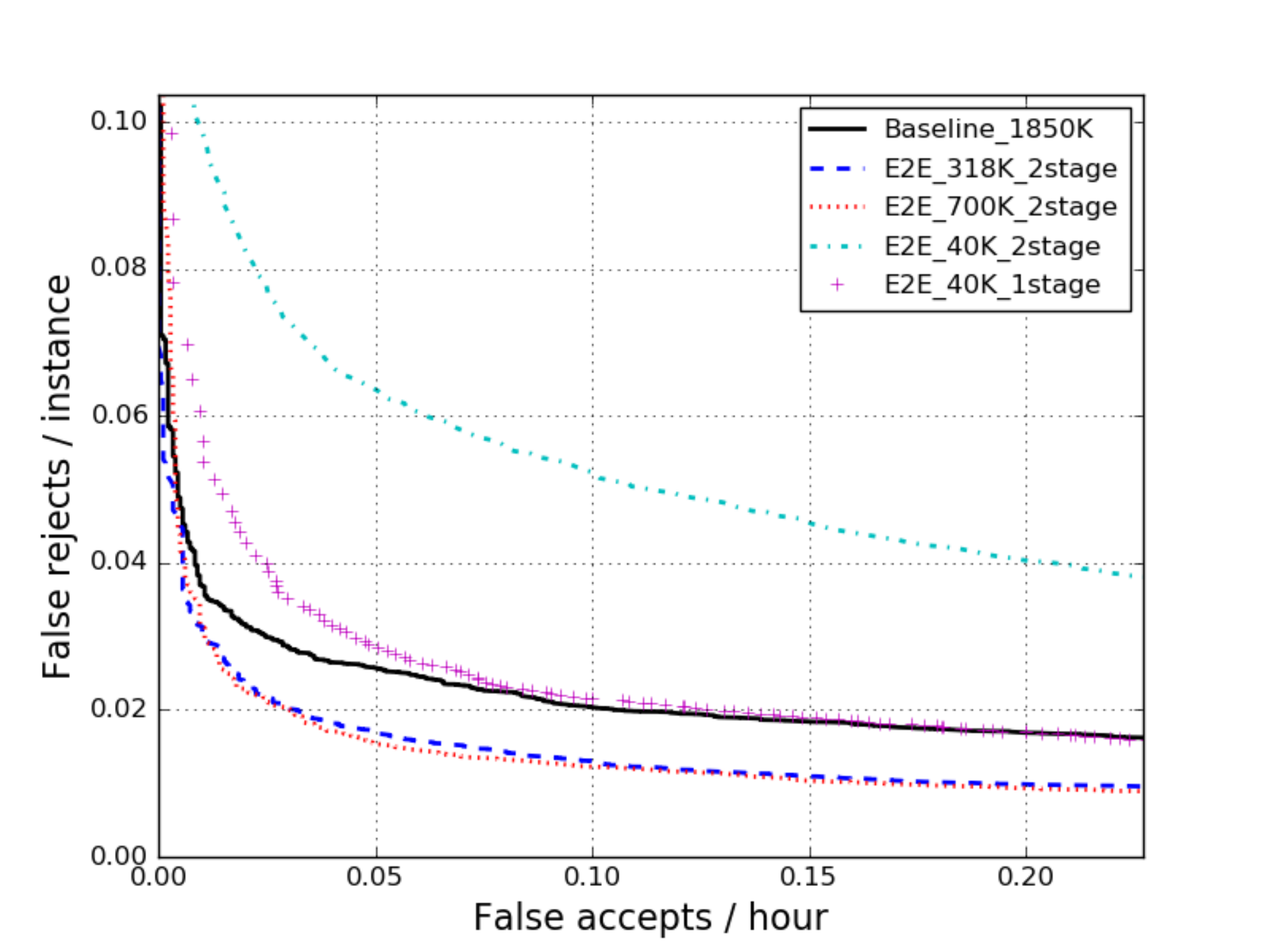}}
		\centerline{(b) Clean accented}\medskip
	\end{minipage}
	\begin{minipage}[b]{\linewidth}
		\centering
		\centerline{\includegraphics[height=137pt]{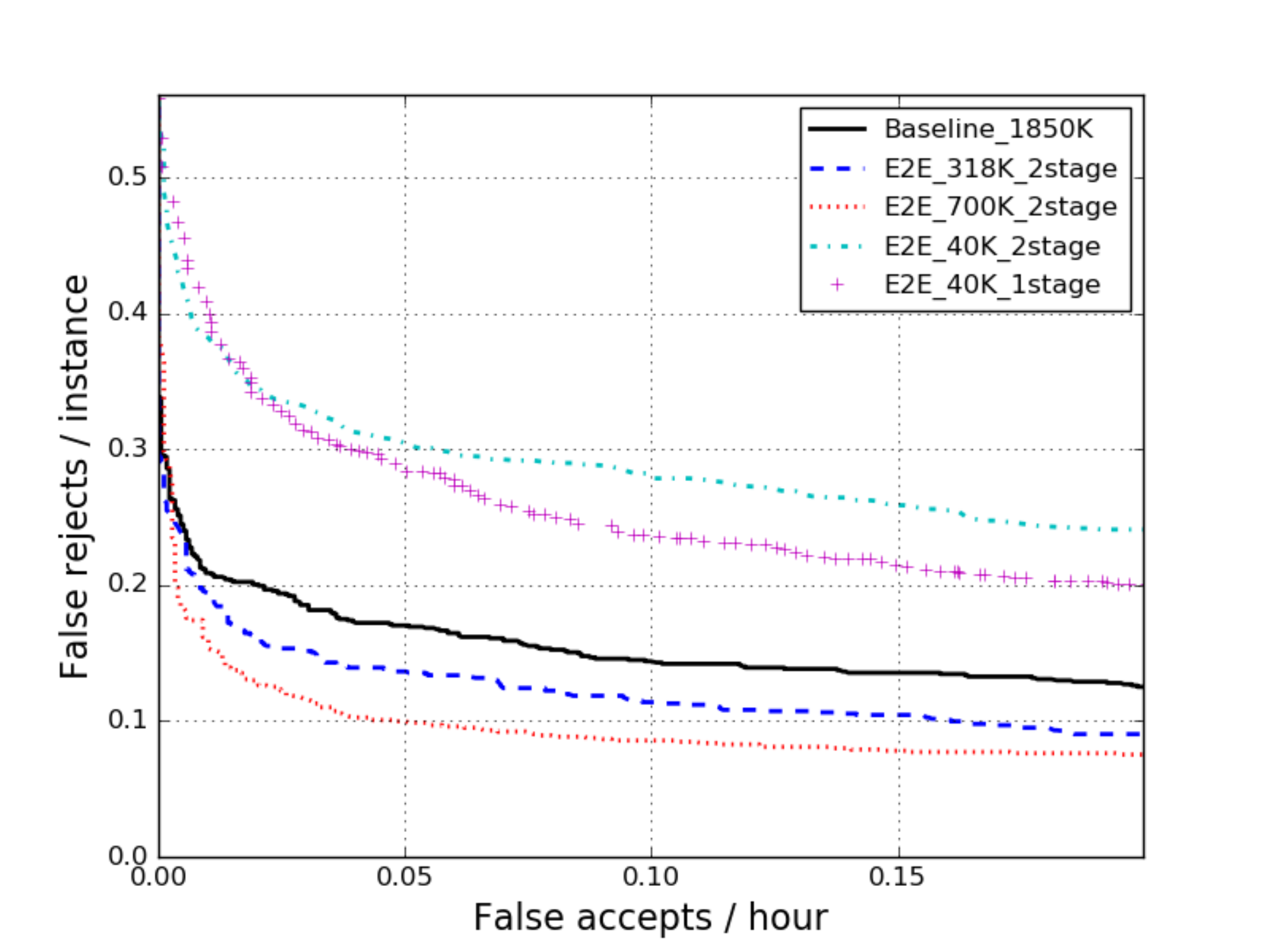}}
		\centerline{(c) High pitched voices}\medskip
	\end{minipage}
	\hfill
	\begin{minipage}[b]{\linewidth}
		\centering
		\centerline{\includegraphics[height=137pt]{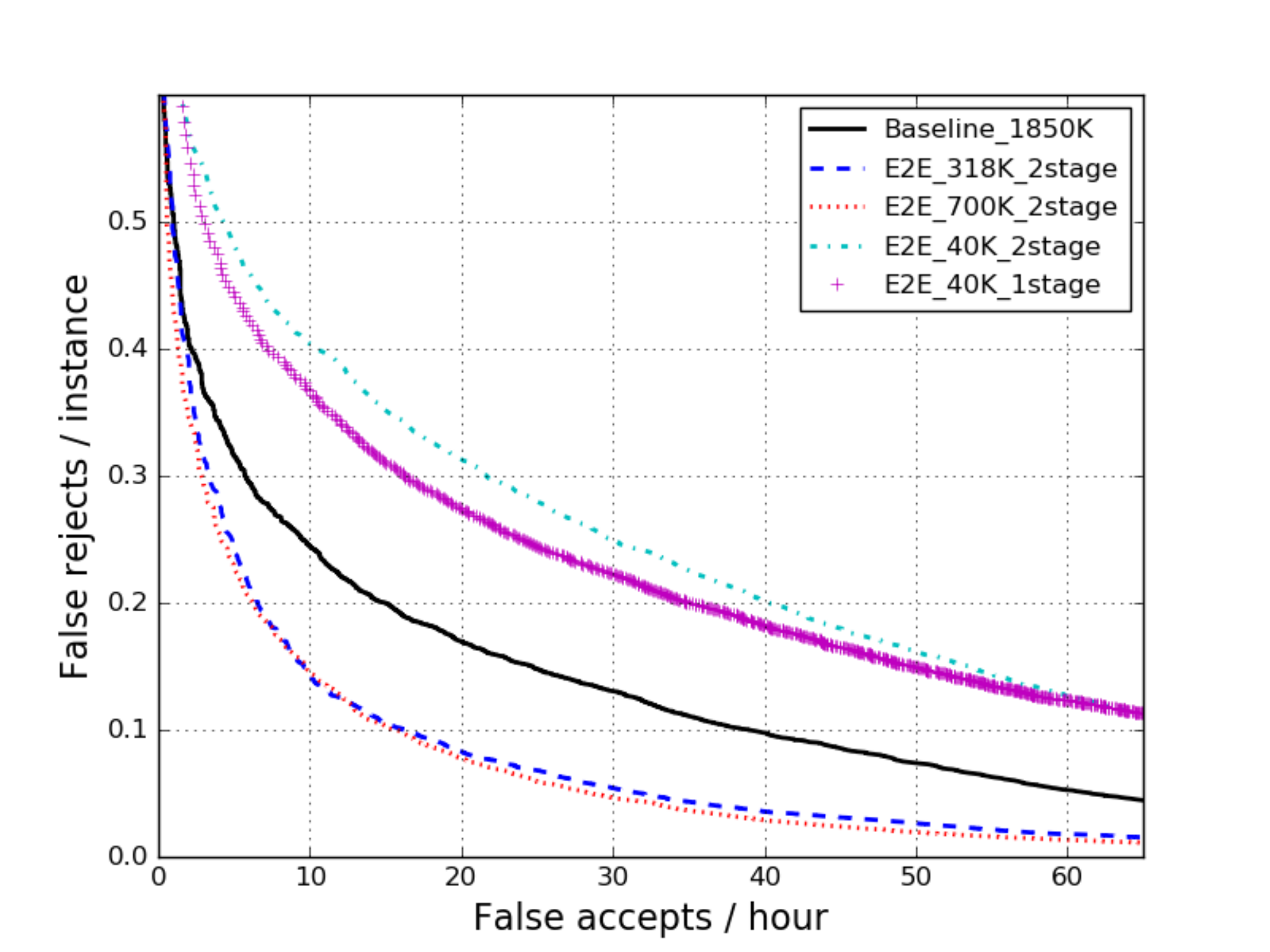}}
		\centerline{(d) Anonymous query logs}\medskip
	\end{minipage}
	\caption{ROC curves under different conditions.}
	\label{fig:roc}
\end{figure}

Our goal is to compare the effectiveness of the proposed approach against the baseline system described in \cite{Cnn15}. Inference is floating point, though (unreported) TensorFlow Lite's quantization~\cite{TfliteSvdfHybridQuant} numbers showed no meaningful degradation. We evaluate the false-reject (FR) and false-accept (FA) tradeoff across several end-to-end models of distinct sizes and computational complexities. As can be seen in the Receiver Operating Characteristic (ROC) curves in Figure \ref{fig:roc}, the 2 largest end-to-end models, with 2-stage training, significantly outperform the recognition quality of the much larger and complex Baseline\_1850K system. More specifically, E2E\_318K\_2stage and E2E\_700K\_2stage show up to 60\% relative FR rate reduction over Baseline\_1850K in most test conditions. Moreover, E2E\_318K\_2stage uses only about 26\% of the computations that Baseline\_1850K uses (once normalizing their execution rates over time), but still shows significant improvements. We also explore end-to-end models at a size that, as described in \cite{Cascade17}, is small enough (in both size and computation) to be executed continuously with very little power consumption. These 2 models, E2E\_40K\_1stage and E2E\_40K\_2stage, also explore the capacity of end-to-end training (\emph{1stage}) versus encoder+decoder training (\emph{2stage}). The ROC curves show that \emph{1stage} training outperforms \emph{2stage} training on all conditions, but particularly on both ``clean" environments where it gets fairly close to the performance of the baseline setup. That is a significant achievement considering E2E\_40K\_1stage has 2.3\% the parameters and performs 3.2\% the computations of Baseline\_1850K. Table \ref{tab:perfdiff} compares the recognition quality of all setups by fixing on a very low false-accept rate of 0.1 FA per hour on a dataset containing only negative (i.e. non-keyword) utterances. Thus the table shows the false-reject rates at that operating point. Here we can appreciate similar trends as those described above: the 2 largest end-to-end models outperform the baseline across all datasets, reducing FR rate about 40\% on the clean conditions, and 40\%-20\% on the other 2 sets depending on the model size. This table also shows how \emph{1stage} outperforms \emph{2stage} for small sized models, and presents similar FR rates as Baseline\_1850K on clean conditions.

\begin{table}[h!]
	\begin{center}
		\caption{FR rate over 4 test conditions at 0.1 FAh level.}
		\label{tab:perfdiff}
		\begin{tabular}{l|r|r|r|r}
			\textbf{Models} & Non Acc. & Accented & High P. & Q. Logs \\
			\hline
			Baseline\_1850K & 1.46\% & 2.03\% & 14.41\% & 12.13\%  \\
			E2E\_318K  & 0.87\% & 1.27\% & 11.39\% &  9.99\%  \\
			E2E\_700K  & 0.87\% & 1.22\% &  8.57\% &  8.90\% \\
			E2E\_40K\_2 & 2.80\% & 5.19\% & 26.54\% & 39.22\%  \\
			E2E\_40K\_1 & 1.52\% & 2.09\% & 23.73\% & 35.32\%  \\
		\end{tabular}
	\end{center}
\end{table}

\section{Conclusion}
\label{sec:conclusion}
We presented a system for keyword spotting that by combining an efficient topology and two types of end-to-end training can significantly outperform previous approaches, at a much lower cost of size and computation. We specifically show how it beats the performance of a setup taken from \cite{Cnn15} with models over 5 times smaller, and even get close to the same performance with a model over 40 times smaller. Our approach provides further benefits of not requiring anything other than a front-end and a neural network to perform the detection, and thus it is easier to extend to newer keywords and/or fine-tune with new training data. Future work includes exploring other loss-functions, as well as generalizing for multi-channel support.

\clearpage
\bibliographystyle{IEEEbib}
\bibliography{strings,refs}

\end{document}